\documentclass[letterpaper]{article}
\usepackage[preprint]{aaai2027}
\usepackage[hyphens]{url}
\usepackage{graphicx}
\usepackage{natbib}
\usepackage{caption}
\usepackage{amsmath}
\usepackage{amssymb}
\usepackage{booktabs}

\input{glyphtounicode}
\pdfglyphtounicode{tfm:wncyr10/S}{0421}
\makeatletter
\DeclareUnicodeCharacter{0421}{%
  {\font\fundingcyr=wncyr10 at \f@size pt\fundingcyr\char83}%
}
\DeclareUnicodeCharacter{2116}{%
  {\font\fundingnumero=ts1-qtmr at \f@size pt\fundingnumero\char155}%
}
\makeatother

\title{Inject Where It Matters: Objective-Dependent Feature Control in Rectified-Flow Image Editing}
\author{Nina Goncharova\textsuperscript{1}, Alexandra Ivanova\textsuperscript{1,2}, Aibek Alanov\textsuperscript{1,2}}
\affiliations{\textsuperscript{1}HSE University\\
\textsuperscript{2}FusionBrain Lab}

\begin{document}

\maketitle

\begin{abstract}
Many prompt-driven FLUX editing methods rely on injecting source features into target-conditioned generation, creating an inherent trade-off between source preservation and the desired edit. We address this problem by controlling where source features are injected and which transformer blocks are used. First, we introduce adaptive masks that spatially gate feature injection using differences between source- and target-conditioned block outputs at matched inversion states. Second, we use REINFORCE to select Top-K injection blocks under different preservation-editability objectives. The selected blocks are fixed at inference and require no online search.
On PIE-Bench with FLUX.1-dev, spatial masking controls the alignment--preservation trade-off. On 60 validation examples, a learned nine-block set gives lower whole-image LPIPS than the 18-block RF-Edit set within the same masked pipeline, with slightly lower alignment. Among 50 random nine-block sets, it has the lowest LPIPS but not the highest reward. Separate objective sweeps reveal different block preferences for preservation and alignment.
\end{abstract}

\section{Introduction}
\label{sec:introduction}

Rectified-flow transformers such as FLUX \citep{liu2023flow,flux2024} define the current state of the art in text-to-image generation, and a growing family of training-free methods adapts them to prompt-driven editing of real images \citep{wang2025taming,deng2025fireflow,Avrahami_2025_CVPR,kim2025reflex}. These methods share a common recipe: the source image is inverted to noise under the source prompt, generation is rerun under the target prompt, and features cached during inversion are injected into the target branch to keep content unrelated to the edit close to the source. This recipe creates an inherent tension: injecting more source information improves preservation but suppresses the requested edit, while injecting less enables the edit at the cost of unrelated changes. How well a method navigates this trade-off is largely determined by two design choices: \emph{where} in the image source features are injected, and \emph{which} transformer blocks are used for injection. We argue that both choices are handled suboptimally by existing methods.

First, injection is spatially uniform. In RF-Edit, FireFlow, and Stable Flow, injected features overwrite every image token of the selected blocks \citep{wang2025taming,deng2025fireflow,Avrahami_2025_CVPR}, so the region that the edit must change receives the same source signal as the background that must stay intact. The only remedy within these methods is to inject less overall, for example at fewer steps or in fewer blocks, which weakens preservation everywhere rather than only where the edit happens. Existing masking strategies do not fill this gap at the feature level. ReFlex derives an editing mask from cross-attention, but applies it through latent blending at early steps, while its feature injection remains spatially uniform \citep{kim2025reflex}. DiffEdit builds its mask from a prompt-difference signal related to ours, but uses one binary region fixed for the whole generation \citep{couairon2023diffedit}.

Second, the injection blocks themselves are chosen with little regard for the editing objective. RF-Edit and FireFlow fix block ranges by hand \citep{wang2025taming,deng2025fireflow}, while Stable Flow selects vital layers through a one-time analysis that removes each layer individually and measures the effect \citep{Avrahami_2025_CVPR}. Such per-layer ablation treats blocks as independent: it does not capture interactions between layers and implicitly assumes that the utility of a layer does not depend on which other layers are selected for injection. Moreover, the resulting set is fixed once and for all, even though different editing objectives, from strict source preservation to strong target alignment, may call for different blocks. Direct search over layer subsets is infeasible: choosing $9$ of the $57$ FLUX.1-dev blocks already yields about $9\times10^9$ candidate sets, each requiring a full editing rollout to evaluate.

We address both limitations. To control where injection happens, we introduce adaptive spatial feature masks built from the token-wise discrepancy between model outputs under the source and target prompts, evaluated at the same states along the inversion trajectory. Tokens where the prompt change matters keep target features, while the remaining tokens receive source features. The masks are recomputed per injection block and per solver step, and require no external segmenter or user-provided region. To decide in which blocks injection is performed, we cast layer selection as policy optimization: REINFORCE \citep{williams1992simple} learns a distribution over $K$-subsets of blocks whose reward is the edit quality itself, a weighted combination of target-prompt alignment and source preservation. Because the policy is rewarded for complete layer sets rather than individual layers, it naturally accounts for interactions between the selected blocks, and changing the reward weighting yields preservation-oriented or alignment-oriented sets from the same procedure. After training, the selected set is fixed and requires no online block search.

Our contributions are as follows:
\begin{itemize}
    \item We propose \textbf{adaptive spatial feature masks} that gate source-feature injection per image token, block, and solver step using prompt-dependent block-output discrepancies at matched inversion states. Controlled analyses on the 140-example PIE-Bench random-editing category examine adaptive versus fixed discrepancy masks and target-retention ratios.
    \item We formulate injection-layer selection as \textbf{reward-driven policy optimization over complete block sets}. On 60 validation examples, the learned Top-$9$ set gives lower LPIPS than our masked 18-block configuration, with slightly lower alignment. It also has lower LPIPS than all 50 sampled random Top-$9$ sets, while 20 random sets attain higher reward. Separate controlled comparisons and cross-procedure transfer use the 140-example analysis subset.
    \item We provide an \textbf{analysis of learned layer preferences} across reward weightings and layer budgets, showing that the optimization objective induces distinct, in the extreme cases disjoint, block selections, and that preservation-focused and alignment-focused objectives favor structurally different parts of the network.
\end{itemize}

\section{Related Work}
\label{sec:related}

\paragraph{Editing with rectified-flow transformers.}
Prompt-driven editing must follow the target prompt while keeping unrelated content intact; with rectified-flow transformers \citep{liu2023flow,flux2024} this is commonly achieved by inverting the source image and transferring source information into target-prompt generation, and methods differ mainly in how much is transferred, where, and through which blocks.
RF-Edit caches self-attention values along RF-Solver's second-order inversion and replays them in selected blocks \citep{wang2025taming}, FireFlow cuts the cost of this scheme with a faster solver \citep{deng2025fireflow}, Stable Flow injects only at vital layers found by ablating individual blocks \citep{Avrahami_2025_CVPR}, ReFlex extracts features at a mid-inversion step and adapts attention during injection \citep{kim2025reflex}, and DNAEdit forgoes injection, aligning the Gaussian noise with the source image in the noise domain \citep{xie2025dnaedit}.
In these injection-based methods the transfer covers every image token of a preselected block set; ReFlex localizes edits through latent blending with a single cross-attention mask, but its feature injection is likewise spatially uniform. We instead gate the transfer per token and learn the blocks.

\paragraph{Localizing edits with masks.}
Spatial masks are the standard tool for deciding \emph{where} an image may change, and they are usually taken from attention maps, as in Prompt-to-Prompt, MasaCtrl, LEDITS++, and FlexiEdit \citep{hertz2023prompttoprompt,cao2023masactrl,brack2024ledits++,koo2024flexiedit}, or from an external segmenter such as SAM \citep{kirillov2023segment}.
DiffEdit instead thresholds the difference between noise predictions under the source and target prompts \citep{couairon2023diffedit}, a signal close to ours, but reduces it to one binary region fixed for the whole generation.
Our masks, built from the token-wise discrepancy between source- and target-conditioned outputs, are recomputed per injection block and solver step and gate feature transfer rather than delimit an editing region.

\paragraph{Choosing injection blocks.}
RF-Edit and FireFlow use predefined block ranges, while Stable Flow ranks blocks through individual removal analyses \citep{wang2025taming,deng2025fireflow,Avrahami_2025_CVPR}. We choose a complete injection set for a specified editing objective using REINFORCE \citep{williams1992simple}, then reuse it across images without online search.
ReCache~\citep{aliev2026recache} similarly uses REINFORCE
to learn discrete controls for diffusion inference,
focusing on caching schedules.

\section{Preliminaries}
\label{sec:preliminaries}

\paragraph{Rectified flow in latent space.}
Let $\mathcal{E}$ and $\mathcal{D}$ be the encoder and decoder of a pretrained
autoencoder. For an image $x$, rectified flow connects
$z^{(0)}=\mathcal{E}(x)$ and $z^{(1)}\sim\mathcal{N}(0,I)$ by linear
interpolation~\citep{liu2023flow}:
\begin{equation*}
    z(\tau)=(1-\tau)z^{(0)}+\tau z^{(1)},
    \qquad \tau\in[0,1].
    \label{eq:rf_path}
\end{equation*}
Its velocity $z^{(1)}-z^{(0)}$ serves as the training target. At inference,
the prompt-conditioned field $v_\phi$ defines
\begin{equation*}
    \frac{d z(\tau)}{d\tau}=v_\phi(z(\tau),\tau,y).
    \label{eq:rf_ode}
\end{equation*}
Here, $\tau=0$ and $\tau=1$ correspond to image and noise latents,
respectively; inversion follows the ODE from image to noise, and generation
in the reverse direction.

For editing, we encode $z_0^*=\mathcal{E}(x_{\mathrm{init}})$ and invert it
under $y_{\mathrm{src}}$ to obtain $z_T^*$. We then set $z_T=z_T^*$, generate
under $y_{\mathrm{trg}}$ down to $\tau=0$, and decode the resulting latent
$z_0$ as $x_{\mathrm{edit}}=\mathcal{D}(z_0)$.

\paragraph{Second-order RF-Solver.}
We apply the second-order RF-Solver~\citep{wang2025taming} to inversion and
generation on a grid $0=\tau_0<\cdots<\tau_T=1$, with
$\Delta_t=\tau_t-\tau_{t-1}>0$ and
$\tau_{t-\frac12}=(\tau_{t-1}+\tau_t)/2$.
A target-generation step from $(z_t,\tau_t)$ to
$(z_{t-1},\tau_{t-1})$ is
\begin{equation*}
\begin{aligned}
    v_t^{(1)}
    &=v_\phi(z_t,\tau_t,y_{\mathrm{trg}}),\\
    \bar z_t
    &=z_t-\frac{\Delta_t}{2}v_t^{(1)},\\
    v_t^{(2)}
    &=v_\phi(\bar z_t,\tau_{t-\frac12},y_{\mathrm{trg}}),\\
    z_{t-1}
    &=z_t-\Delta_t v_t^{(2)}.
\end{aligned}
\end{equation*}
Although only $v_t^{(2)}$ appears in the final update, $v_t^{(1)}$
determines its evaluation point; hence, both evaluations affect the step.
Inversion applies the same midpoint scheme in the forward-time direction:
\begin{align*}
    \bar z_t^*
    &=
    z_{t-1}^*
    +
    \frac{\Delta_t}{2}
    v_\phi(z_{t-1}^*,\tau_{t-1},y_{\mathrm{src}}),\\
    z_t^*
    &=
    z_{t-1}^*
    +
    \Delta_t
    v_\phi(\bar z_t^*,\tau_{t-\frac12},y_{\mathrm{src}}).
\end{align*}

\section{Method}
\label{sec:method}

We keep the pretrained velocity field and RF-Solver fixed and modify target
generation only through source-feature injection. Adaptive spatial masks
determine at which image tokens source features are injected, while a learned
policy selects the transformer blocks in which injection is performed.

\subsection{Setting and Notation}
\label{sec:setting}

\paragraph{Image-editing task.}
Given a source image $x_{\mathrm{init}}$, its source prompt
$y_{\mathrm{src}}$, and a target prompt $y_{\mathrm{trg}}$, our goal is to
produce an edited image $x_{\mathrm{edit}}$ that follows
$y_{\mathrm{trg}}$ while preserving source content unrelated to the prompt
change.

\paragraph{Source and target branches.}
Using the grid introduced in Section~\ref{sec:preliminaries}, inversion under
$y_{\mathrm{src}}$ produces the source trajectory
$\{z_t^*\}_{t=0}^{T}$, where $z_t^*$ is the state at flow time $\tau_t$.
Target generation under $y_{\mathrm{trg}}$ starts from $z_T=z_T^*$ and
produces the target trajectory $\{z_t\}_{t=0}^{T}$, where $z_t$ is the state
at flow time $\tau_t$. We refer to these trajectories as the source and
target branches.

Injection is applied during the first $N$ target-generation steps,
\[
    z_t\to z_{t-1},
    \qquad
    t\in\mathcal{I}_N=\{T-N+1,\ldots,T\},
\]
where $N\leq T$.

\paragraph{Model and injection layers.}
We use a frozen FLUX.1-dev model~\citep{flux2024} as the rectified-flow backbone. The model contains $L=57$ transformer
blocks: 19 double-stream blocks followed by 38 single-stream blocks. We refer
to each transformer block as an injection layer and allow feature injection at
any of the $L$ layers. We define
\[
    \mathcal{L}=\{0,\ldots,L-1\},
    \qquad
    \mathcal{S}_K=\{S\subseteq\mathcal{L}:|S|=K\},
\]
where $S\in\mathcal{S}_K$ is a set of $K$ injection layers. During target
generation, the spatially masked feature injection described in
Section~\ref{sec:masks} is applied only at layers $\ell\in S$; all remaining
layers are evaluated without modification.

\paragraph{Solver evaluations and branch features.}
For each target-generation step $t\in\mathcal{I}_N$, let $r=1$ and $r=2$
denote the current-state and predicted-midpoint evaluations introduced in
Section~\ref{sec:preliminaries}. Their source- and target-branch states are
\[
    \begin{array}{c|cc}
        & \text{source branch} & \text{target branch} \\
        \hline
        r=1
        & (z_t^*,\tau_t)
        & (z_t,\tau_t) \\
        r=2
        & (\bar z_t^*,\tau_{t-\frac{1}{2}})
        & (\bar z_t,\tau_{t-\frac{1}{2}}).
    \end{array}
\]

For $\ell\in S$, let
\[
    h_{\ell,t,r}^{*},\,h_{\ell,t,r}\in\mathbb{R}^{P\times D}
\]
denote the corresponding source- and target-branch image-token outputs.
Here, $P$ is the number of image tokens and $D=3072$; the batch dimension
is omitted.

During inversion, $h_{\ell,t,2}^*$ is cached at the midpoint of
$[\tau_{t-1},\tau_t]$. For $t\in\mathcal{I}_N$ with $t<T$,
$h_{\ell,t,1}^*$ is reused from the current-state evaluation of the next
interval, which starts at $z_t^*$. Since no subsequent interval starts at
$z_T^*$, we explicitly evaluate $z_T^*$ at $\tau_T$ under
$y_{\mathrm{src}}$ to obtain $h_{\ell,T,1}^*$.

Following RF-Edit~\citep{wang2025taming}, we use these source outputs in the corresponding target
evaluations. Unlike RF-Edit, which injects attention value tensors, we inject
block outputs: the updated image representation in double-stream blocks and
the image-token portion of the joint branch output in single-stream blocks.
We compare output and value injection in Supplementary
Section~\ref{sec:supp_output_value}.

\subsection{Adaptive Spatial Feature Masks}
\label{sec:masks}

For each $t\in\mathcal{I}_N$ and $\ell\in S$, we construct an endpoint mask
$M_{\ell,t}$ from layer outputs evaluated at $z_t^*$, the endpoint of the
inversion interval $[\tau_{t-1},\tau_t]$. Let
$h_{\ell,t,1}^{*}(y)$ denote the output obtained by evaluating $z_t^*$ under
prompt $y$ at flow time $\tau_t$. We cache
$h_{\ell,t,1}^{*}(y_{\mathrm{src}})=h_{\ell,t,1}^{*}$ for injection and also
evaluate the same state under $y_{\mathrm{trg}}$ to construct the mask.

For each image token $n$, we compute the channel-averaged squared discrepancy
\begin{equation}
    d_{\ell,t,n}
    =
    \frac{1}{D}
    \left\lVert
        h_{\ell,t,1,n}^{*}(y_{\mathrm{trg}})
        -
        h_{\ell,t,1,n}^{*}(y_{\mathrm{src}})
    \right\rVert_2^2.
    \label{eq:endpoint_discrepancy}
\end{equation}
Because the two outputs are evaluated at the same latent, flow time, and
guidance, $d_{\ell,t,n}$ measures the effect of changing the prompt.

For a target-retention ratio $\rho\in(0,1]$, let
$k_\rho=\max\{1,\operatorname{round}(\rho P)\}$. We set
$M_{\ell,t,n}=1$ at the $k_\rho$ tokens with the largest discrepancies and
$M_{\ell,t,n}=0$ elsewhere. For each solver evaluation $r\in\{1,2\}$, the
masked layer output is
\begin{equation}
    \widetilde{h}_{\ell,t,r}
    =
    M_{\ell,t}\odot h_{\ell,t,r}
    +
    (1-M_{\ell,t})\odot h_{\ell,t,r}^{*},
    \label{eq:masked_injection}
\end{equation}
where the mask is broadcast across feature channels. Thus, tokens with
$M_{\ell,t,n}=1$ retain target features, while the remaining tokens receive
source features. Setting $M_{\ell,t}=0$ gives spatially uniform image-token injection under this formulation; $\rho=1$ retains target features everywhere. Our legacy unmasked controls additionally differ in single-stream text-token handling and, in the mask-construction run, source-cache alignment, so they are not an exact implementation of this image-token-only limit.

The same mask is used for both solver evaluations. Injection during the first
affects the predicted midpoint, while injection during the second affects the
full-step update.

\begin{figure}[t]
    \centering
    \includegraphics[width=1\linewidth]{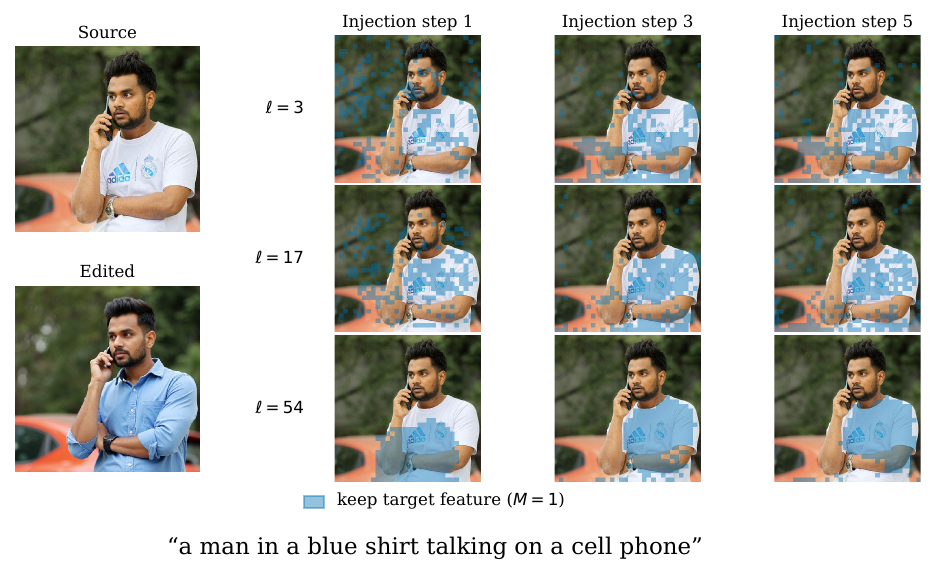}
    \caption{Adaptive endpoint masks across three injection layers and steps.
Blue tokens retain target features ($M_{\ell,t,n}=1$); source features are
injected elsewhere.}
    \label{fig:adaptive_mask}
\end{figure}

In double-stream blocks, Equation~\eqref{eq:masked_injection} is applied to the image-stream output. In single-stream blocks, it is applied only to the
image-token portion of the joint output, leaving text tokens unchanged. Masks
are constructed online without an external segmenter or a user-provided
spatial mask. Figure~\ref{fig:adaptive_mask} shows the resulting endpoint
masks at three layers and three injection steps.

\subsection{Policy-Based Layer Selection}
\label{sec:policy}

For an injection-layer set $S\in\mathcal{S}_K$, let
$x_{\mathrm{edit}}(S)$ denote the edited image obtained using the masked
injection procedure described above, with all other settings held fixed.

Selecting $K$ injection layers from $L$ candidates defines a search space of
size
\[
    |\mathcal{S}_K|=\binom{L}{K}.
\]
For FLUX.1-dev, selecting $K=9$ layers among $L=57$ candidates yields
approximately $9\times10^9$ possible sets. Evaluating each set requires a
target-generation rollout and metric computation, making exhaustive search
infeasible. We therefore learn a distribution over layer sets using
REINFORCE~\citep{williams1992simple}.

We assign a learnable parameter $\theta_\ell$ to each candidate layer $\ell\in\mathcal{L}$ and
define
\begin{equation*}
    q_\theta(\ell)
    =
    \frac{\exp(-\theta_\ell)}
    {\sum_{j\in\mathcal{L}}\exp(-\theta_j)}.
    \label{eq:layer_probability}
\end{equation*}
Here, $q_\theta(\ell)$ is the probability of selecting layer $\ell$ at the
first sampling step; smaller $\theta_\ell$ corresponds to higher probability. The
policy is global and is not conditioned on the image, prompts, source
features, or masks.

To select exactly $K$ distinct layers, we sample an ordered sequence
$a=(\ell_1,\ldots,\ell_K)$ without replacement:
\begin{equation}
    \pi_\theta(a)
    =
    \prod_{k=1}^{K}
    \frac{\exp(-\theta_{\ell_k})}
    {\sum_{j\in\mathcal{L}\setminus
    \{\ell_1,\ldots,\ell_{k-1}\}}
    \exp(-\theta_j)}.
    \label{eq:ordered_policy}
\end{equation}
After each selection, the chosen layer is removed from the remaining pool.
Although $\pi_\theta$ is defined over ordered sequences, the editing procedure
uses only the set $S(a)=\{\ell_1,\ldots,\ell_K\}$, so its reward is independent
of the sampled order.

For each training example, we define the reward of an injection-layer set $S$ using
CLIPScore~\citep{hessel2021clipscore} and
LPIPS~\citep{zhang2018unreasonable} as
\begin{equation}
    \begin{aligned}
        R_\alpha(S)
        ={}&
        \alpha\,
        \operatorname{CLIPScore}
        \left(x_{\mathrm{edit}}(S),y_{\mathrm{trg}}\right)
        \\
        &-
        \operatorname{LPIPS}
        \left(x_{\mathrm{edit}}(S),x_{\mathrm{init}}\right).
    \end{aligned}
    \label{eq:reward}
\end{equation}
The coefficient $\alpha\geq0$ balances target-prompt alignment against source
preservation. Larger $\alpha$ favors alignment, while LPIPS penalizes source
changes. CLIP-only policies omit the LPIPS term.

Let $\mathcal{D}_{\mathrm{train}}$ denote the empirical distribution of
training examples. We maximize
\begin{equation}
    J(\theta)
    =
    \mathbb{E}_{\substack{
        (x_{\mathrm{init}},y_{\mathrm{src}},y_{\mathrm{trg}})
        \sim\mathcal{D}_{\mathrm{train}},\\
        a\sim\pi_\theta
    }}
    \left[
        R_\alpha(S(a))
    \right].
    \label{eq:policy_objective}
\end{equation}
We treat the editing rollout and metric computations as black-box evaluations
and optimize $J(\theta)$ with REINFORCE.

For each training example, we sample $m\geq 2$ sequences
$a_1,\ldots,a_m\sim\pi_\theta$. Generation randomness is held fixed across
these rollouts, so only the sampled injection-layer set varies. For rollout
$i$, we use the leave-one-out baseline
\begin{equation*}
    b_i
    =
    \frac{1}{m-1}
    \sum_{\substack{j=1\\j\neq i}}^{m}
    R_\alpha(S(a_j)),
    \label{eq:loo_baseline}
\end{equation*}
and define the advantage
\[
    A_i = R_\alpha(S(a_i))-b_i.
\]
The resulting Monte Carlo policy-gradient estimator is
\begin{equation*}
    \widehat{\nabla_\theta J}
    =
    \frac{1}{m}
    \sum_{i=1}^{m}
    A_i\,\nabla_\theta \log \pi_\theta(a_i).
    \label{eq:policy_gradient}
\end{equation*}
We implement gradient ascent on $J$ by minimizing
\begin{equation}
    \mathcal{L}_{\mathrm{policy}}
    =
    -\frac{1}{m}
    \sum_{i=1}^{m}
    A_i\,\log \pi_\theta(a_i).
    \label{eq:policy_loss}
\end{equation}
The advantages are treated as constants during differentiation. Thus,
gradients propagate only through $\log\pi_\theta(a_i)$, and only $\theta$ is
optimized; the generative model and metric networks remain frozen.

After training, we form a deterministic set from the $K$ layers with the
largest values of $q_\theta(\ell)$, equivalently the $K$ smallest values of
$\theta_\ell$. The set is fixed at inference and requires neither sampling nor
online search. Additional training and implementation details are provided in
Supplementary Section~\ref{sec:supp_policy}.

\subsection{Analyzing Injection-Layer Preferences}
\label{sec:policy_analysis}

We use policies trained on real PIE-Bench images~\citep{ju2024pnpinversion} to study how injection-layer
preferences depend on the editing objective and the number of selected layers $K$.

\paragraph{Real-image policy sweeps.}
For each pair $(\alpha,K)$, we train an independent policy using real-image
source inversion and endpoint-masked target generation. Across the sweep, we
use the same training examples, solver, number of injection steps, injected
feature type, masking rule, and rollout budget; only the reward weight
$\alpha$ and layer budget $K$ vary. We denote the learned probabilities of
selecting each layer at the first sampling step by
\[
    q_{\alpha,K}(\ell)
    =
    q_{\theta^{(\alpha,K)}}(\ell).
\]
Smaller values of $\alpha$ emphasize source preservation through the LPIPS
term, while larger values place more weight on target-prompt alignment.
Because CLIPScore and LPIPS have different numerical scales, $\alpha$ is
interpreted relative to the fixed metric implementations used throughout our
experiments.

\paragraph{Preference maps.}
For each $(\alpha,K)$, we rank all $L$ layers by
$q_{\alpha,K}(\ell)$ and visualize the ranks as a heatmap, with rank one
denoting the most preferred layer. Because layers are sampled sequentially
without replacement, these are ranks of the first-step selection
probabilities, not marginal probabilities of inclusion in the final
$K$-layer set. Section~\ref{sec:block_analysis_results} presents the resulting
maps and their analysis.

\section{Experiments}
\label{sec:experiments}

We evaluate adaptive spatial masking and policy-based layer selection with
FLUX.1-dev on real PIE-Bench images. We report controlled comparisons and
cross-procedure transfer, analyze learned layer preferences, and compare our
complete method with FLUX-based editing baselines.

\subsection{Experimental Setup}
\label{sec:experimental_setup}

\paragraph{Data.}
We shuffle the ordered 700-example PIE-Bench mapping~\citep{ju2024pnpinversion} with Python's \texttt{random.Random(42)}. The first 640 examples form the policy-optimization split and the remaining 60 form the validation split. End-to-end evaluation and the 50-random-set comparison use all 60 validation examples, covering all ten editing types. These examples are excluded from policy updates. A fixed 20-example subset was monitored during optimization, so this validation split is not an untouched test set.

Controlled layer-set comparisons, mask ablations, objective sweeps, cross-procedure transfer, and qualitative examples use the fixed 140-example random-editing category (type 0) in public mapping-file order. This category overlaps the optimization and validation splits by 124 and 16 examples, respectively. We report these controlled analyses separately from validation performance.

\paragraph{Editing protocol.}
Controlled experiments use the same nonuniform 25-step FLUX schedule for inversion and
generation, with guidance scales $1.0$ and $3.5$, respectively. Unless stated
otherwise, we use $K=9$, $N=5$, and $\rho=0.2$. Trade-off curves vary
$N\in\{0,2,5,10\}$, where $N=0$ denotes no feature injection.

\paragraph{Policy optimization.}
We train the real-image policy with a 30-step RF-Solver schedule using
$\alpha=0.15$, $K=9$, and $m=8$ sampled sets per example. Optimization uses a
learning rate of $0.06$, no entropy regularization, and gradient accumulation
over eight examples. After 160 optimizer steps, corresponding to two passes over the 640-example optimization split and 10,240 editing rollouts, we extract the deterministic Top-$9$ set from the final policy checkpoint. This set is used in our layer-set, transfer, and end-to-end
evaluations. Our RF-Solver evaluations use 25 steps; transfer retains
each procedure's native solver schedule.

\paragraph{Metrics.}
We report one alignment and one preservation metric.
CLIPScore~\citep{hessel2021clipscore} scores target-prompt alignment without
requiring a ground-truth edit, which real images lack, and
LPIPS~\citep{zhang2018unreasonable} measures perceptual deviation from the
source, including both intended edits and unrelated changes. Neither
whole-image metric isolates local edit correctness. The same metrics define the reward in
Equation~\eqref{eq:reward}. We compute CLIPScore with the ViT-L/14 model
\texttt{RE-N-Y/clipscore-vit-large-patch14} and LPIPS with an AlexNet
backbone, both on the full image.

\paragraph{Comparisons.}
To isolate the effect of layer selection, we evaluate the layer sets used by
RF-Edit and Stable Flow, random sets, and our learned set within the same
RF-Solver pipeline. We test transferability by integrating the learned set into
the RF-Edit and FireFlow pipelines. End-to-end comparisons retain each baseline's native solver and injection
configuration while matching target guidance at 3.5. Metrics use
$512\times512$ outputs; Stable Flow generates at $1024\times1024$
and is bicubically downsampled before evaluation.

\subsection{Adaptive Spatial Masks}
\label{sec:mask_experiments}

We compare adaptive and fixed discrepancy masks using the 18-block RF-Edit set on the 140-example analysis category. Both conditions share text-token handling and endpoint source-feature alignment. At $N=5$, adaptive masks give CLIPScore/LPIPS of $26.6378/0.4504$, versus $26.3595/0.4460$ for a per-image mask fixed at block 54 and step index 4. Adaptation increases alignment with slightly greater perceptual change. Varying $\rho$ provides another way to choose the trade-off. Legacy unmasked and random-mask controls also differ in implementation details, as specified in Supplementary Section~\ref{sec:supp_masks}.

\begin{figure*}[t]
    \centering
    \includegraphics[width=0.85\linewidth]{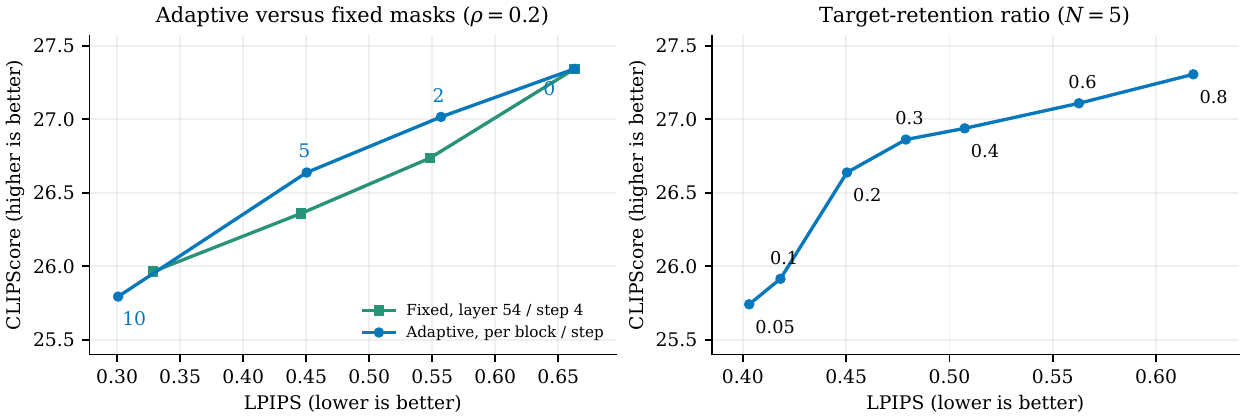}
    \caption{Spatial control on 140 random-editing examples with the 18-block RF-Edit set. Left: adaptive and fixed endpoint masks at $\rho=0.2$, with labels $N\in\{0,2,5,10\}$. Right: varying $\rho$ at $N=5$. All displayed conditions share text-token handling and endpoint source features.}
    \label{fig:mask_pareto}
\end{figure*}

\subsection{Controlled Layer-Set Comparison}
\label{sec:layer_selection_results}

\begin{figure}[t]
    \centering
    \includegraphics[width=0.75\linewidth]{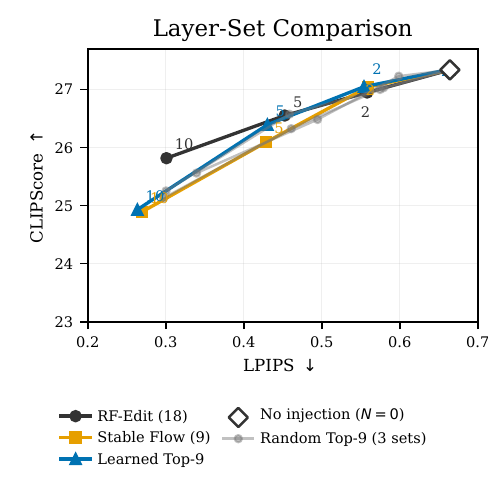}
    \caption{Layer-set trade-offs on the 140-example analysis category across $N$; learned uses step 160.}
    \label{fig:layers_pareto}
\end{figure}

We isolate layer selection on the 140-example analysis category by holding
the 25-step endpoint-masked RF-Solver pipeline fixed. At the matched budget $K=9$, we compare the Stable Flow set,
three random sets sampled uniformly without replacement from 57 blocks, and
our Top-$9$ set obtained from the step-160 policy checkpoint trained with
$\alpha=0.15$. The 18-layer RF-Edit set serves as a higher-budget reference.
Table~\ref{tab:layer_selection} reports results at $N=5$, while
Figure~\ref{fig:layers_pareto} varies $N$.

\begin{table}[t]
    \centering
    {\small
    \setlength{\tabcolsep}{1mm}
    \begin{tabular}{@{}lcrrr@{}}
        \toprule
        Layer set
        & $K$
        & \shortstack{CLIPScore\\$\uparrow$}
        & \shortstack{LPIPS\\$\downarrow$}
        & \shortstack{Reward\\$\uparrow$} \\
        \midrule
        Stable Flow
        & 9 & 26.092 & 0.429 & 3.485 \\

        Random
        & 9 & 26.460 & 0.471 & 3.498 \\
        \multicolumn{2}{@{}l}{\quad $\pm$ sample s.d.}
        & 0.126 & 0.020 & 0.026 \\

        Learned (ours)
        & 9 & 26.392 & 0.430 & \textbf{3.529} \\
        \midrule
        RF-Edit
        & 18 & 26.550 & 0.453 & 3.530 \\
        \bottomrule
    \end{tabular}
    }
    \caption{Layer-set comparison at $N=5$. RF-Edit is a higher-budget
    reference; bold marks the best reward among the $K=9$ sets.
    Random results are computed across three sets; reward uses
    $\alpha=0.15$. Results use the 140-example analysis category
    (124 optimization and 16 validation examples); the separate 50-set
    validation comparison is in Section~\ref{sec:random50}.}
    \label{tab:layer_selection}
\end{table}

At $N=5$, the learned set improves CLIPScore over Stable
Flow~\citep{Avrahami_2025_CVPR} at nearly identical LPIPS, and matches the
18-layer RF-Edit reward~\citep{wang2025taming} at half the budget with lower
LPIPS. Its reward leads the $K=9$ sets, though this is the objective it
optimized, and exceeds the random mean by about one sample s.d. The CLIPScore
lead over Stable Flow holds at all nonzero $N$, despite training at $N=5$.

\subsection{Cross-Procedure Transfer of Layer Sets}
\label{sec:layer_transfer}

This transfer analysis uses the 140-example category, not the validation split.

We test whether layer preferences transfer across editing procedures.
Section~\ref{sec:layer_selection_results} evaluates the RF-Edit and Stable
Flow layer sets~\citep{wang2025taming,Avrahami_2025_CVPR} in our
endpoint-masked RF-Solver pipeline.

Conversely, we replace the native layer sets of RF-Edit and FireFlow~\citep{deng2025fireflow} with the
same learned Top-$9$ set extracted from the step-160 policy checkpoint, while
retaining their respective solvers, schedules, and native feature-injection
mechanisms; our endpoint masks are not used. This reduces the injection budget from 18 to 9 blocks in RF-Edit and from 38
to 9 in FireFlow, while preserving the CLIPScore--LPIPS trade-off in both
procedures and slightly improving it for some FireFlow configurations. Supporting transfer results are provided in
Supplementary Section~\ref{sec:supp_layers}.

\subsection{Learned Injection-Layer Preferences}
\label{sec:block_analysis_results}

All metrics in this objective sweep use the 140-example analysis category. Each objective--budget configuration has one training run; stability across training seeds remains unmeasured.

We train real-image policies for $K\in\{3,6,9\}$ with reward weights
$\alpha\in\{0,0.01,0.15,0.4,1\}$ and an additional CLIP-only objective. For a matched comparison across the full sweep, all preference maps use the
common checkpoint available after 80 optimizer steps.

\begin{figure}[t]
    \centering
    \includegraphics[width=\linewidth]{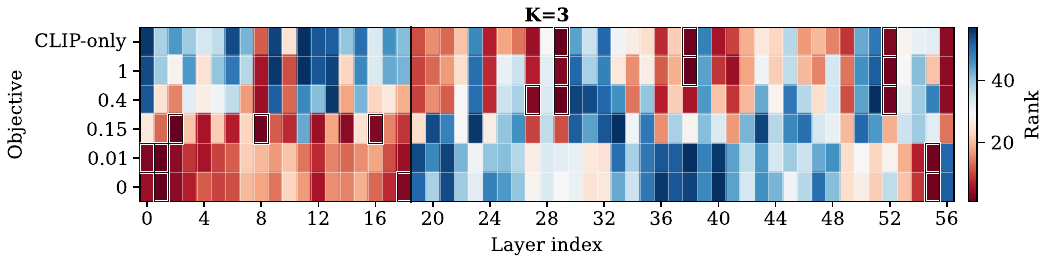}
    \includegraphics[width=\linewidth]{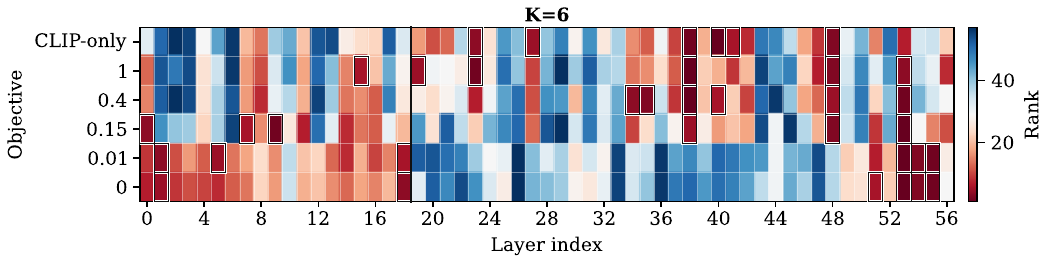}
    \includegraphics[width=\linewidth]{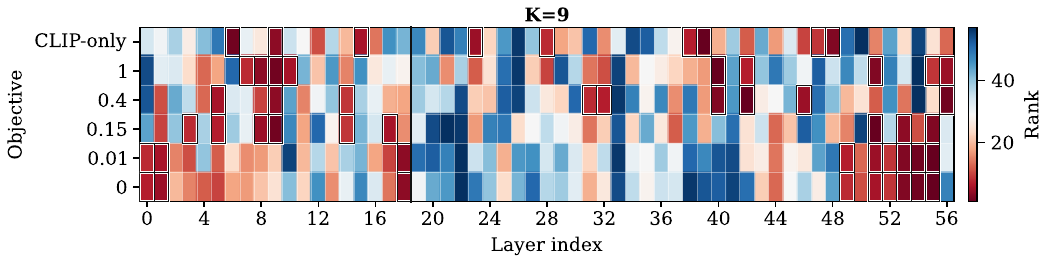}
    \caption{Layer rankings across objectives for $K\in\{3,6,9\}$. Red indicates
higher preference. Blocks 0--18/19--56 are double-/single-stream; all policies
use step 80.}
    \label{fig:preference_maps}
\end{figure}

Figure~\ref{fig:preference_maps} shows that layer preferences depend on both
the objective and the layer budget. At the two objective extremes, the
LPIPS-only ($\alpha=0$) and CLIP-only Top-$K$ sets are disjoint for every
$K\in\{3,6,9\}$.

\begin{figure*}[t]
    \centering
    \includegraphics[width=\textwidth]{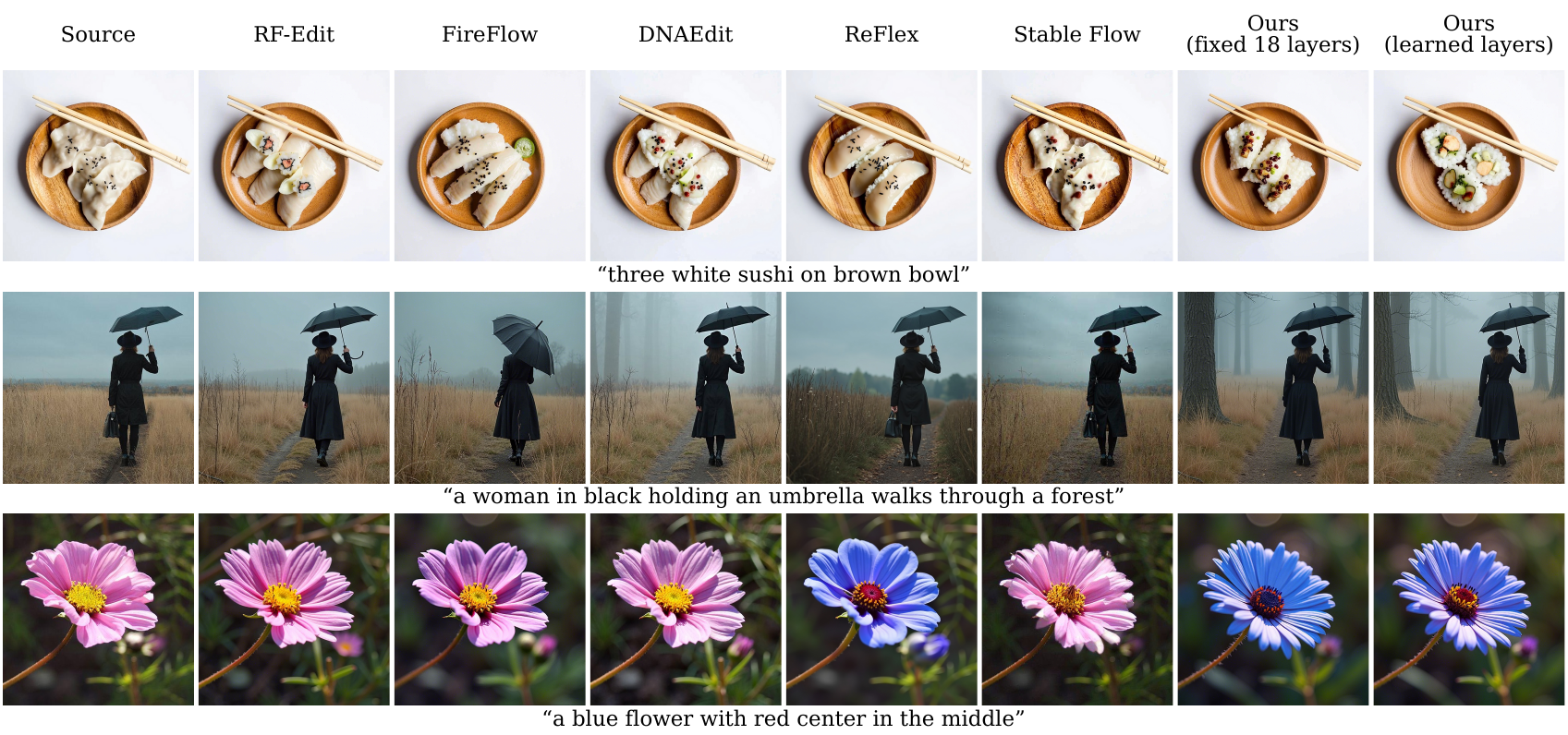}
    \caption{Qualitative comparison on the 140-example PIE-Bench analysis category. Each row shows
    the source image, baseline edits, and our outputs using the RF-Edit and
    learned layer sets; the target prompt is shown below.}
    \label{fig:editing_results}
\end{figure*}

The LPIPS-only sets are strictly nested across budgets,
$S_3\subset S_6\subset S_9$: preservation-focused optimization concentrates
on a compact core of blocks that is retained as the budget grows, combining
early and final double-stream blocks with late single-stream blocks. In contrast, the
CLIP-only sets for $K=3$ and $K=6$ contain only middle and late single-stream
blocks. The CLIP-only Top-$9$ set additionally contains three double-stream
blocks and is not a nested extension of the smaller sets.

Controlled evaluations confirm this distinction. At $K=9$ and $N=5$, the
LPIPS-only set attains an LPIPS of $0.420$, compared with $0.482$ for the
CLIP-only set, whereas the CLIP-only set increases CLIPScore from $26.054$
to $26.556$. The same pattern holds across all nine evaluated $(K,N)$ pairs,
showing that different objectives induce distinct layer preferences and
corresponding alignment--preservation trade-offs. Supporting results are
provided in Supplementary Section~\ref{sec:supp_layers}.

\subsection{End-to-End Evaluation}
\label{sec:end_to_end}

\begin{figure}[t]
    \centering
    \includegraphics[width=0.9\linewidth]{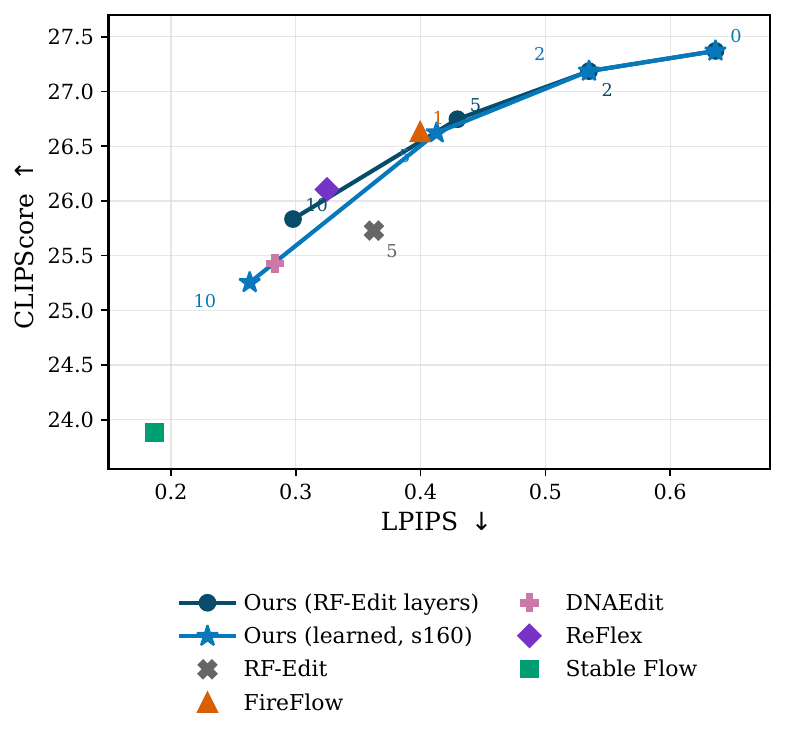}
    \caption{Whole-image CLIPScore--LPIPS trade-offs on 60 validation examples. Labels denote $N$; $N=0$ is shared by our two variants. Native baseline points use the configurations in Table~\ref{tab:supp_method_configs}, including RF-Edit at $N=5$ and FireFlow at $N=1$. Target guidance is 3.5. Upper left is better; connecting lines only guide the eye.}
    \label{fig:end_to_end_tradeoff}
\end{figure}

We compare our full pipeline with five FLUX-based baselines on the 60-example validation split. The methods retain their native solver and injection configurations; target guidance is 3.5 for all methods. Table~\ref{tab:end_to_end} reports one operating point per method, and Figure~\ref{fig:end_to_end_tradeoff} varies the injection horizon only for our two configurations.

\begin{table}[t]
    \centering
    {\small
    \setlength{\tabcolsep}{1mm}
    \begin{tabular}{@{}lcrr@{}}
        \toprule
        Method & Blocks & CLIPScore $\uparrow$ & LPIPS $\downarrow$ \\
        \midrule
        RF-Edit & 18 & 25.728 & 0.3627 \\
        FireFlow & 38 & 26.634 & 0.3997 \\
        DNAEdit & -- & 25.428 & 0.2835 \\
        ReFlex & 32 & 26.105 & 0.3250 \\
        Stable Flow & 9 & 23.881 & \textbf{0.1867} \\
        \midrule
        Ours (RF-Edit layers) & 18 & \textbf{26.747} & 0.4295 \\
        Ours (learned layers) & 9 & 26.623 & 0.4129 \\
        \bottomrule
    \end{tabular}
    }
    \caption{End-to-end results on 60 validation examples with FLUX.1-dev. Ours and RF-Edit use $N=5$; FireFlow uses $N=1$. Target guidance is 3.5. Our variants differ only in injection-block set. Counts describe injection/sharing blocks, not executed backbone blocks. Column bests are bold.}
    \label{tab:end_to_end}
\end{table}

At $N=5$, our learned Top-$9$ set gives lower LPIPS than our 18-block variant ($0.4129$ versus $0.4295$), with slightly lower CLIPScore ($26.623$ versus $26.747$). Native FireFlow has both higher CLIPScore and lower LPIPS than our learned point at these settings. Stable Flow has the lowest LPIPS and lowest CLIPScore.

At $N=2$, our variants agree at the reported precision ($27.187/0.535$). At $N=10$, our 18-block variant gives $25.834/0.2978$: it has higher alignment but greater source deviation than DNAEdit ($25.428/0.2835$), and lower alignment and source deviation than ReFlex ($26.105/0.3250$). The learned variant gives $25.253/0.2631$. These are trade-offs, not dominance over both baselines. The horizon sweep adds sampled nondominated operating points; it is not an exhaustive or interpolated baseline comparison.

\paragraph{Comparison with random block sets.}
On the same validation split, our learned Top-$9$ set has lower LPIPS than all 50 sampled random nine-block sets and is Pareto-nondominated among them. Its reward is close to the random mean ($3.5805$ versus $3.5782$), and 20 random sets have higher reward. Supplementary Section~\ref{sec:random50} gives the matched protocol and full comparison.

Qualitative examples in Figure~\ref{fig:editing_results} and Supplementary Figure~\ref{fig:supp_qualitative_comparison} come from the separate 140-example runs. Whole-image LPIPS includes intended edits, so it cannot establish that the requested edit was completed.

\paragraph{Inference cost.}
All backbone blocks still execute. The injection-block count does not measure total compute. At $N=5$ with 25 inversion and generation steps on the 140-example protocol, our masked variants use 110 transformer calls per edit versus 100 for native RF-Edit. End-to-end latency and peak memory remain unmeasured.

\section{Conclusion}
\label{sec:conclusion}

We studied where source features are injected and which transformer blocks carry them in rectified-flow image editing. Adaptive spatial masks and objective-dependent block selection provide controls over the alignment--preservation trade-off. On 60 validation examples, nine learned injection blocks yield lower whole-image LPIPS than our masked 18-block configuration, with slightly lower alignment. The learned set has the lowest LPIPS among 50 sampled random sets, but not the highest reward. Separate 140-example analyses show objective-dependent block preferences and cross-procedure transfer. These results concern the reported operating points and whole-image metrics; they do not establish an inference speedup or uniformly better local edit correctness.

\section*{Acknowledgments}
We thank Mishan Aliev for helpful consultations on stabilizing REINFORCE training and for his mere existence.

The work was supported by the grant for research centers in the field of AI provided by the Ministry of Economic Development of the Russian Federation in accordance with the agreement 000000С313925P4E0002 and the agreement with HSE University № 139-15-2025-009. This research was supported in part through computational resources of HPC facilities at HSE University.

\bibliography{aaai2027}

\clearpage
\appendix

\section*{Supplementary Material}

\section{Additional Ablations}
\label{sec:supp_ablations}

The quantitative ablations in this section use the 140-example analysis category, including 124 policy-optimization examples. They are separate from the 60-example validation comparison.

\subsection{Comparison of Output and Value Injection}
\label{sec:supp_output_value}

We compare the attention-value injection used by
RF-Edit~\citep{wang2025taming} with the block-output injection used in our
method. We disable spatial masking and hold the solver, examples,
generation seeds, and layer set fixed. Legacy single-stream output injection
also replaces text-token outputs, so this comparison does not isolate
representation type alone. The learned comparison uses the step-80 Top-$9$ set.

With the RF-Edit layer set, value and output injection produce similar
trade-offs. At $N=5$, value injection gives a CLIPScore of $25.502$ and LPIPS
of $0.382$, while output injection gives $25.554$ and $0.390$, respectively.
For the learned set, output injection reduces LPIPS from $0.524$ to $0.408$,
while CLIPScore decreases from $26.792$ to $25.814$. Thus, the injected
feature changes the alignment--preservation trade-off, but neither type
uniformly dominates the other.

\subsection{Adaptive Spatial Masks}
\label{sec:supp_masks}

We vary the target-retention ratio and compare adaptive endpoint masks with
fixed and matched-area random masks. These experiments use a separate
140-example ablation run exported on July 25, 2026, with real-image inversion
and the 18-layer RF-Edit set. We retain this run's own controls and do not
merge its values with the controlled layer-set comparison.

\begin{figure*}[t]
    \centering
    \includegraphics[width=\linewidth]{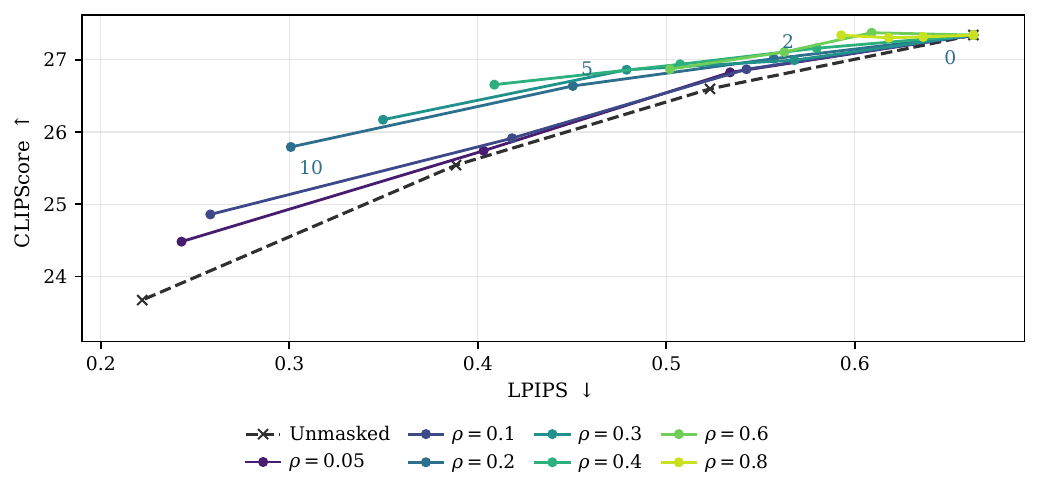}
    \caption{Effect of the target-retention ratio $\rho$. Curves vary
    $N\in\{0,2,5,10\}$; labels denote $N$ for the default $\rho=0.2$ curve.
    The dashed curve shows this separate run's legacy unmasked
    output-injection control, which also differs in text-token handling
    and first-order source-cache alignment.}
    \label{fig:supp_mask_ratio}
\end{figure*}

\paragraph{Target-retention ratio.}
The target-retention ratio $\rho$ controls the fraction of image tokens that
retain target features. Figure~\ref{fig:supp_mask_ratio} varies
$\rho\in\{0.05,0.1,0.2,0.3,0.4,0.6,0.8\}$ while holding the solver, layer set,
examples, and generation seeds fixed. Increasing $\rho$ generally shifts the
result toward higher CLIPScore and higher LPIPS, providing direct control over
the alignment--preservation trade-off. At $N=5$, increasing $\rho$ from
$0.05$ to $0.8$ changes CLIPScore from $25.741$ to $27.305$ and LPIPS from
$0.403$ to $0.618$. We use $\rho=0.2$ in the main experiments as an
intermediate setting.

\paragraph{Mask construction.}
The separate mask-ablation run evaluates 12 conditions at
$N\in\{0,2,5,10\}$ on 140 examples: a legacy unmasked control, seven
adaptive retention ratios, one fixed discrepancy mask, and three random-mask
seeds. Every condition has 140 metric records at each $N$. The no-injection
outputs agree across all conditions, with CLIPScore $27.3428$ and LPIPS
$0.6629$; their per-example metric differences are zero.

At $\rho=0.2$, we compare masks recomputed per block and step with a
per-image discrepancy mask from layer 54 and step index 4, reused for all
injections. These adaptive and fixed conditions share text-token handling
and endpoint source-feature caching, isolating mask reuse versus adaptation.
Random masks retain exactly the same fraction of image tokens and are fixed
across blocks and steps for each image, with seeds 42, 98, and 791. They share
the masked text-token rule. However, the available ablation runner refreshes
the first-order endpoint cache for adaptive and fixed discrepancy masks,
but not for random masks or the legacy unmasked control. Those two controls
therefore provide comparisons between procedures that also differ in
source-feature alignment.

\begin{figure}[!t]
    \centering
    \includegraphics[width=\linewidth]{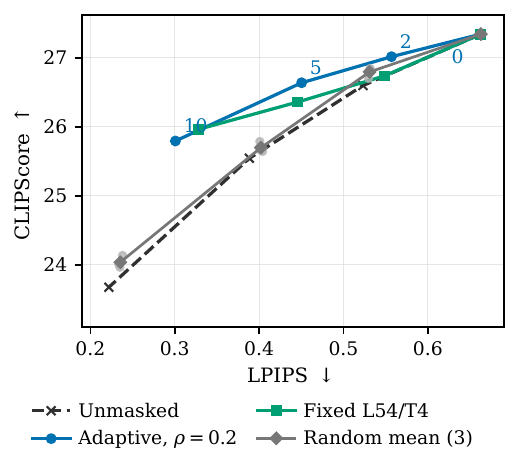}
    \caption{Mask constructions in the separate 140-example ablation run
    exported on July 25, 2026, using the 18-layer RF-Edit set and
    $\rho=0.2$. Labels denote $N$. Gray diamonds average three fixed random
    masks, with individual seeds shown faintly. The dashed curve is this
    run's own legacy unmasked control. Adaptive and fixed discrepancy masks
    share text-token handling and endpoint caches; random and unmasked
    controls retain the earlier cache alignment.}
    \label{fig:supp_mask_constructions}
\end{figure}

At $N=5$, adaptive and fixed discrepancy masks give
$(26.6378,0.4504)$ and $(26.3595,0.4460)$ in CLIPScore--LPIPS,
respectively: the adaptive mask has higher mean alignment with slightly
greater perceptual change. At $N=10$, the adaptive mask instead gives lower
LPIPS and lower alignment, $(25.7938,0.3008)$ versus $(25.9643,0.3286)$.
Thus adaptation changes the operating point without uniformly improving
both metrics. The random-mask mean at $N=5$ is $(25.6977,0.4021)$, and the
legacy unmasked control gives $(25.5429,0.3885)$. Their lower alignment and
lower perceptual change provide additional context for the sampled
trade-offs in Figure~\ref{fig:supp_mask_constructions}.

\begin{figure*}[p]
    \centering
    \includegraphics[width=\textwidth,height=0.84\textheight,keepaspectratio]{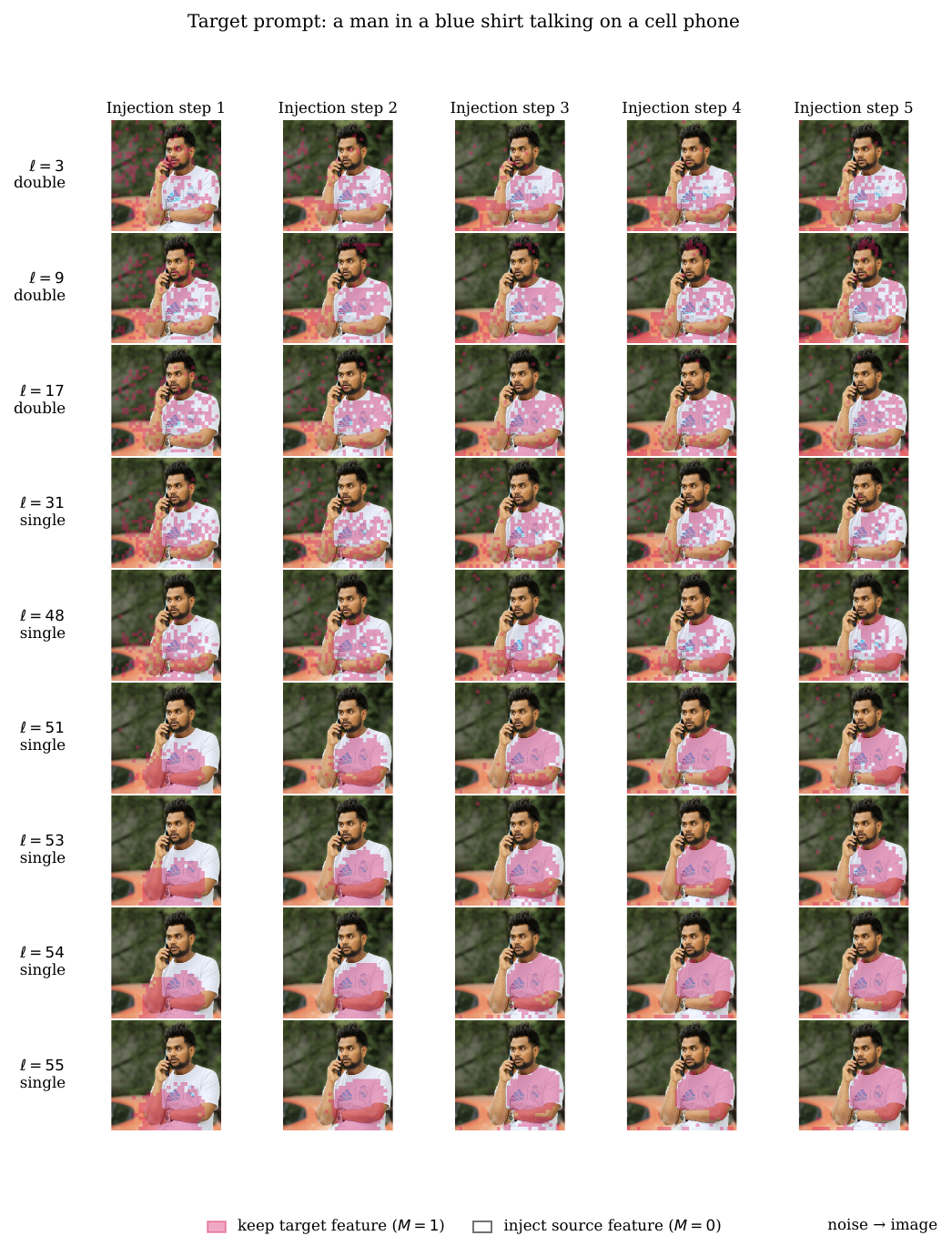}
    \caption{Endpoint masks for the final step-160 Top-$9$ set across five
    injection steps. Highlighted tokens retain target features.}
    \label{fig:supp_mask_grid}
\end{figure*}

\paragraph{Masks across layers and steps.}
Figure~\ref{fig:supp_mask_grid} shows that the endpoint masks for the final
step-160 Top-$9$ set vary across both injection layers and steps.

\begin{figure*}[t]
    \centering
    \includegraphics[width=\textwidth]{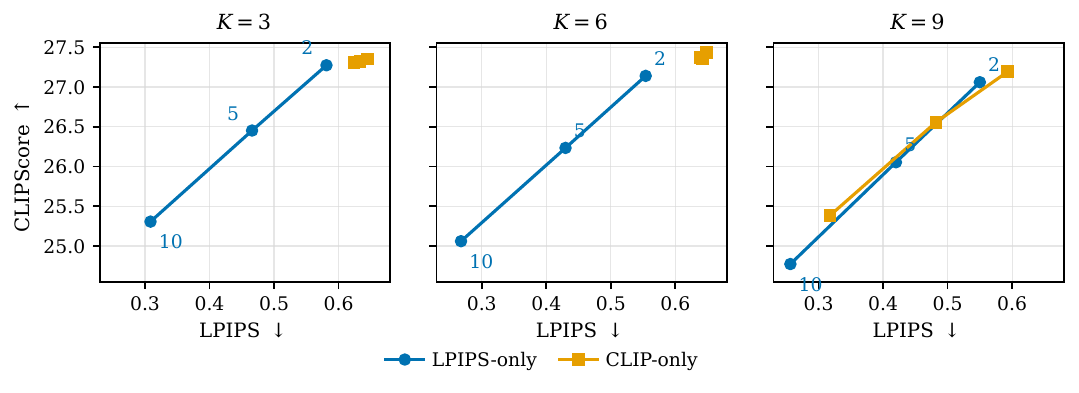}
    \caption{CLIPScore--LPIPS trade-offs for LPIPS-only and CLIP-only
    policies at $K\in\{3,6,9\}$. Labels on the LPIPS-only curves denote
    $N\in\{2,5,10\}$; the CLIP-only curves use the same values. Upper left
    is better.}
    \label{fig:supp_extreme_objectives}
\end{figure*}

\section{Policy Optimization Details}
\label{sec:supp_policy}

\paragraph{Algorithmic outline.}
For each training image--prompt triplet, we (1) encode and invert the source
image while caching source-branch block outputs, (2) construct endpoint masks
from source- and target-conditioned outputs at the same inversion states,
(3) sample $m$ ordered $K$-block subsets from the policy, (4) generate one
edit per subset with masked injection and a shared generation seed, (5)
compute CLIPScore--LPIPS rewards, and (6) form prompt-wise leave-one-out
advantages and update only the policy parameters. The generative model,
autoencoder, and metric networks remain frozen throughout this procedure.

\paragraph{Subset sampling.}
The policy contains one scalar parameter $\theta_\ell$ for each of the 57
candidate FLUX.1-dev blocks. We implement the ordered sampling distribution in
Equation~\eqref{eq:ordered_policy} using equivalent weighted sampling without
replacement. We compute the log probability of the sampled order by
successively masking selected blocks, while the editing procedure receives
only the resulting unordered block set.

\paragraph{Training batches and baseline.}
For each example, we evaluate $m=8$ layer sets with the same generation seed
and compute the leave-one-out baseline within this group. We accumulate
gradients over eight examples, yielding 64 editing rollouts per update.
Advantages are used without normalization or clipping.

\begin{table}[t]
    \centering
    \normalsize
    \setlength{\tabcolsep}{5pt}
    \renewcommand{\arraystretch}{1.05}
    \begin{tabular}{@{}ll@{}}
        \toprule
        Parameter & Value \\
        \midrule
        Candidate blocks & 57 \\
        Sampled blocks $K$ & 9 (main policy) \\
        Sets per example $m$ & 8 \\
        Examples per optimizer update & 8 \\
        Editing rollouts per update & 64 \\
        Optimizer & Adam \\
        Learning rate & $0.06$ \\
        Weight decay & $0$ \\
        Entropy coefficient & $0$ \\
        Logit initialization & $\mathcal{N}(1,0.03^2)$ \\
        RF-Solver steps during training & 30 \\
        Injection steps during training & $N=5$ \\
        Target-retention ratio & $\rho=0.2$ \\
        Inversion / generation guidance & $1.0$ / $3.5$ \\
        \bottomrule
    \end{tabular}
    \caption{Policy-optimization settings shared by the reported real-image
    runs. Only $K$ and the reward objective vary in the preference sweep.}
    \label{tab:supp_policy_settings}
\end{table}

\paragraph{Development ranges and selection.}
Preliminary development, including experiments preceding the final
endpoint-mask formulation, explored 13 learning rates in $[0.001,0.1]$,
rollout-group sizes $m\in\{1,2,4,8,16\}$, accumulation factors $\{4,8\}$,
entropy coefficients $\{0,0.01\}$, initialization standard deviations
$\{0,0.03\}$, layer budgets $K\in\{3,6,9,12,18\}$, training horizons
$\{80,160,320,640\}$, and injection horizons $N\in\{5,10\}$.
Guided by stable preliminary optimization and the available rollout budget, we
fixed the settings in Table~\ref{tab:supp_policy_settings} before the
objective--budget sweep, which trains $K\in\{3,6,9\}$ against
$\alpha\in\{0,0.01,0.1,0.15,0.4,1\}$ and a CLIP-only objective; $K=9$ matches
the Stable Flow budget, and $\alpha=0.15$ and $\rho=0.2$ are intermediate
points of their sweeps.
Figure~\ref{fig:preference_maps} shows every objective except the additional
$\alpha=0.1$ development run.

\paragraph{Optimization and checkpointing.}
Only the 57 policy parameters are optimized; FLUX.1-dev, the autoencoder, and
the metric networks remain frozen. Checkpoints are stored every 40 optimizer
steps, and we run one policy-optimization trial per objective--budget
configuration.
Each of the 21 configurations is trained for 80 optimizer steps: one pass over
the 640-example optimization split, or $80\times64=5{,}120$ rollouts. The
primary $\alpha=0.15$, $K=9$ policy continues to 160 steps (two passes,
$10{,}240$ rollouts), and its final step-160 checkpoint is used for the
controlled, transfer, and end-to-end evaluations.

\paragraph{Validation.}
We monitor optimization on a fixed 20-example subset of the 60-example
validation split, evaluated before training and every 20 optimizer steps.
Neither these monitoring results nor the controlled analyses are used for
checkpoint selection. We subsequently evaluate end-to-end methods and 50
random block sets on all 60 validation examples. The split is excluded
from policy updates but was not untouched during development.

\paragraph{Data splits.}
We construct the policy split by shuffling the ordered PIE-Bench mapping with
Python's \texttt{random.Random(42)}: the first 640 examples form the
optimization split and the next 60 form the validation split. End-to-end
and 50-random-set evaluations use these same 60 IDs. Controlled analyses
retain the 140 examples with editing-type identifier zero in the original
mapping-file order; 124 belong to optimization and 16 to validation.

\paragraph{Randomness.}
We use seed 42 for the PIE-Bench split, policy initialization, and generation.
Before each rollout, Python, NumPy, PyTorch, and CUDA random-number generators
are seeded explicitly. Let $g$ denote the zero-indexed rollout group, which
advances once for each image--prompt micro-batch. The $i$-th policy draw in
that group uses
\[
    42 + 8g + i,
    \qquad i\in\{0,\ldots,7\}.
\]
Thus, each accumulated example receives a distinct group of eight policy
seeds. Generation is then reset to seed 42, so the eight rollouts entering a
leave-one-out comparison differ only in the sampled layer set. Training
examples are reshuffled at epoch $e$ using seed $42+e$. Reported evaluation
curves likewise use generation seed 42 for every compared configuration.

The controlled random baseline uses the following three independently sampled
9-layer sets:
\[
\begin{aligned}
S_{\mathrm{rand},1}&=\{8,10,19,20,23,24,35,39,46\},\\
S_{\mathrm{rand},2}&=\{9,18,20,22,26,30,31,36,53\},\\
S_{\mathrm{rand},3}&=\{2,11,12,16,17,22,29,42,52\}.
\end{aligned}
\]
These realized sets are reused for every injection schedule. Likewise, the
three realized matched-area random masks are fixed across comparisons; neither
baseline is resampled for different schedules.

\paragraph{Computing infrastructure.}
Policy optimization and evaluation were run on single NVIDIA A100 or H100
GPUs with 80\,GB of memory. A typical 80-step policy-optimization run took
approximately 13 hours. The allocated containers exposed 2--12 CPU cores on
Intel Xeon Gold 6278C hosts and used 96--243\,GB of host memory.

Our RF-Solver and policy jobs used Ubuntu 20.04 in the container image
\texttt{cuda12.1-torch2-py311:0.0.36}. Recorded runtimes report
\texttt{Python 3.11}, NVIDIA driver \texttt{560.35.03},
\texttt{CUDA 12.6}, \texttt{PyTorch 2.10.0+cu126},
\texttt{Transformers 5.3.0}, \texttt{lpips 0.1.4}, and
\texttt{imscore 0.0.11}. Our RF-Solver implementation uses the original FLUX
code and does not depend on Diffusers.

Native Stable Flow used \texttt{PyTorch 2.4.1+cu121} and
\texttt{Diffusers 0.30.0}. The DNAEdit and ReFlex environments used
\texttt{PyTorch 2.4.1+cu121} and \texttt{Diffusers 0.31.0}; ReFlex additionally
used \texttt{Transformers 4.44.2}. Our RF-Solver inference used bfloat16 parameters,
while policy parameters, rewards, and solver accumulations used float32.

\section{Additional Layer-Selection Results}
\label{sec:supp_layers}

Objective, checkpoint, and transfer metrics in this section use the 140-example analysis category. Section~\ref{sec:random50} separately reports the 50-set validation comparison.

\paragraph{Reference layer sets.}
Using the unified indexing of the main paper, the 18-block RF-Edit set is
$\{39,\ldots,56\}$, and the 9-block Stable Flow set is
$\{0,1,17,18,25,28,53,54,56\}$. These fixed sets are used whenever the
corresponding layer configuration is evaluated in our RF-Solver pipeline.

\begin{figure*}[t]
    \centering
    \includegraphics[width=\linewidth]{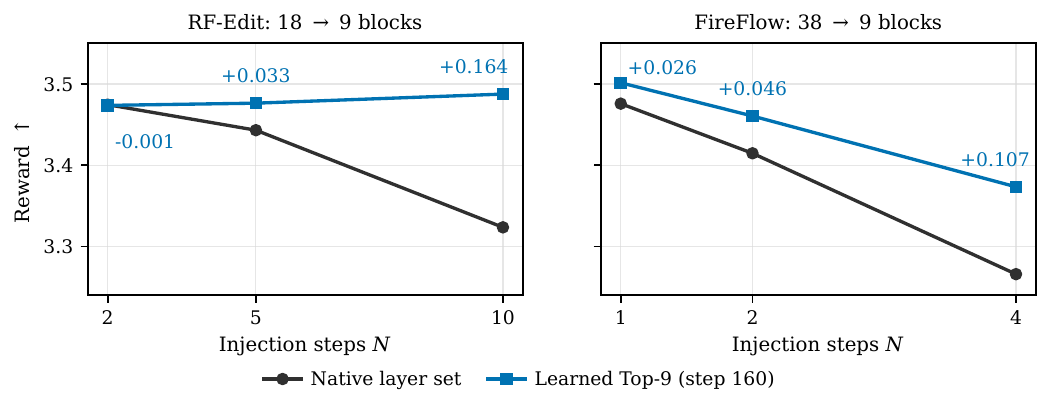}
    \caption{Cross-procedure transfer of the final step-160 Top-$9$ set to
    RF-Edit (left, $18\to9$ blocks) and FireFlow (right, $38\to9$ blocks).
    Labels show the change in $R_{0.15}$ relative to the native layer set at
    the same $N$.}
    \label{fig:supp_transfer}
\end{figure*}

\subsection{Layer Preferences Across Objectives and Budgets}
\label{sec:supp_objectives}

We evaluate policies trained with $K\in\{3,6,9\}$ and objectives
$\alpha\in\{0,0.01,0.1,0.15,0.4,1\}$ together with a CLIP-only objective, all
taken from the step-80 checkpoint for comparability and evaluated with endpoint
masks; unless noted, we report the $N=5$ operating point.

The two extremes are especially distinct: the LPIPS-only and CLIP-only
Top-$K$ sets are disjoint at every evaluated budget
(Table~\ref{tab:supp_extreme_layer_sets}), and the LPIPS-only sets are nested,
$S_3\subset S_6\subset S_9$, whereas the CLIP-only sets change more
substantially with $K$.

\begin{table}[t]
    \centering
    \small
    \setlength{\tabcolsep}{3pt}
    \renewcommand{\arraystretch}{1.08}
    \begin{tabular}{@{}cll@{}}
        \toprule
        $K$ & LPIPS-only Top-$K$ & CLIP-only Top-$K$ \\
        \midrule
        3 & $\{1,18,55\}$ & $\{29,38,52\}$ \\
        \addlinespace[3pt]
        6 & \shortstack[c]{$\{1,18,51,$\\$53,54,55\}$}
          & \shortstack[c]{$\{23,27,38,$\\$40,41,48\}$} \\
        \addlinespace[3pt]
        9 & \shortstack[c]{$\{0,1,18,49,51,$\\$52,53,54,55\}$}
          & \shortstack[c]{$\{6,9,15,23,28,$\\$38,39,47,48\}$} \\
        \bottomrule
    \end{tabular}
    \caption{Deterministic Top-$K$ layer sets learned with the LPIPS-only and
    CLIP-only objectives. Layer indices follow the unified FLUX transformer
    indexing used in the main paper.}
    \label{tab:supp_extreme_layer_sets}
\end{table}

At $K=9$ and $N=5$, LPIPS-only gives $(26.054,0.420)$ in
CLIPScore--LPIPS, whereas CLIP-only gives $(26.556,0.482)$.
Figure~\ref{fig:supp_extreme_objectives} shows the same preservation--alignment
shift across all nine evaluated $(K,N)$ pairs.

\subsection{Step-80 and Step-160 Layer Sets}
\label{sec:supp_checkpoints}

For a matched training budget, we evaluate every $(\alpha,K)$ configuration
after 80 optimizer steps. We then continue the final $\alpha=0.15$, $K=9$
policy to 160 steps. Table~\ref{tab:supp_checkpoint_comparison} compares its
step-80 and step-160 layer sets and metrics.

\begin{table}[t]
    \centering
    \small
    \setlength{\tabcolsep}{5pt}
    \begin{tabular}{@{}lcc@{}}
        \toprule
        & Step 80 & Step 160 \\
        \midrule
        \raisebox{0.55\baselineskip}{Top-$9$ set}
        & \shortstack{$\{3,5,8,9,14,$\\$17,51,53,55\}$}
        & \shortstack{$\{3,9,17,31,48,$\\$51,53,54,55\}$} \\
        CLIPScore $\uparrow$ & 26.368 & \textbf{26.392} \\
        LPIPS $\downarrow$ & 0.440 & \textbf{0.430} \\
        Reward $\uparrow$ & 3.515 & \textbf{3.529} \\
        \bottomrule
    \end{tabular}
    \caption{Step-80 and step-160 checkpoints for $\alpha=0.15$ and $K=9$
    under the endpoint-masked protocol at $N=5$. Reward is
    $0.15\,\mathrm{CLIPScore}-\mathrm{LPIPS}$.}
    \label{tab:supp_checkpoint_comparison}
\end{table}

All objective--budget comparisons in
Section~\ref{sec:supp_objectives} use step 80. The step-160 set slightly
improves both metrics and is used in the final controlled, transfer, and
end-to-end evaluations.

\subsection{Cross-Procedure Transfer}
\label{sec:supp_transfer}

We transfer the final step-160 Top-$9$ set to RF-Edit and FireFlow while
retaining each procedure's native solver, schedule, and feature-injection
mechanism; endpoint masks are disabled. We compare layer sets using
\[
    R_{0.15}=0.15\,\mathrm{CLIPScore}-\mathrm{LPIPS}.
\]

In RF-Edit, replacing 18 native blocks with 9 learned blocks changes reward by
$-0.001$, $+0.033$, and $+0.164$ for $N\in\{2,5,10\}$, respectively
(Figure~\ref{fig:supp_transfer}, left). Thus, the learned set approximately
matches the native set at $N=2$ and improves reward at the two longer injection
schedules.

In FireFlow, replacing 38 native blocks with 9 learned blocks improves reward
by $+0.026$, $+0.046$, and $+0.107$ for $N\in\{1,2,4\}$, respectively
(Figure~\ref{fig:supp_transfer}, right). Overall, the transferred set improves
reward in five of the six evaluated configurations and is within $0.001$ of the
native set in the sixth, while substantially reducing the injection budget.

\subsection{Comparison with 50 random block sets}
\label{sec:random50}

We compare the learned step-160 Top-$9$ set with 50 distinct random sets of
nine blocks on the 60-example validation split. Each set is evaluated on all
60 examples using the same endpoint-masked output-injection pipeline:
25 inversion and generation steps, $N=5$, $\rho=0.2$, target guidance 3.5,
and generation seed 42. All $50\times60=3{,}000$ random-set edits completed
successfully. These results use validation examples across all ten editing
types and are separate from the three-set controlled comparison in
Table~\ref{tab:layer_selection}.

\begin{table}[t]
    \centering
    \small
    \setlength{\tabcolsep}{2pt}
    \caption{Nine-block sets on the 60-example validation split at $N=5$.
    Each set's metrics are averaged over the same 60 examples; the first two rows
    report the mean and sample standard deviation across 50 sets.
    The highest-reward random set is candidate 08. Reward is
    $0.15\,\mathrm{CLIPScore}-\mathrm{LPIPS}$.}
    \label{tab:random50}
    \begin{tabular}{@{}lccc@{}}
        \toprule
        Layer set & CLIPScore $\uparrow$ & LPIPS $\downarrow$
        & $R_{0.15}$ $\uparrow$ \\
        \midrule
        Random (mean)
        & 26.841 & 0.4480 & 3.5782 \\
        $\quad\pm$ sample s.d.
        & 0.209 & 0.0135 & 0.0273 \\
        Random (highest reward)
        & 27.228 & 0.4427 & 3.6414 \\
        Learned (step 160)
        & 26.623 & 0.4129 & 3.5805 \\
        \bottomrule
    \end{tabular}

\end{table}

\begin{figure}[!htbp]
    \centering
    \includegraphics[width=0.90\linewidth]{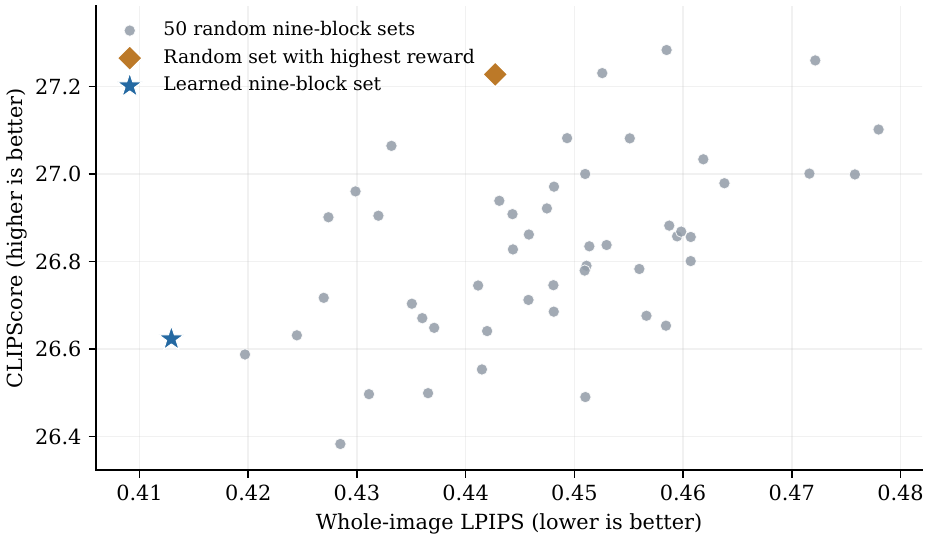}
    \caption{The learned Top-$9$ set and 50 random nine-block sets under the
    same endpoint-masked protocol on 60 validation examples. Each point
    averages one fixed set over the full split. Higher CLIPScore and lower
    LPIPS are preferred.}
    \label{fig:random50}
\end{figure}

The learned set has lower LPIPS than every sampled random set: $0.4129$
versus a random-set minimum of $0.4197$. Its CLIPScore is $0.218$ below the
random mean, with LPIPS lower by $0.0350$. Consequently its reward is close
to that mean ($3.5805$ versus $3.5782$), and 20 of the 50 random sets attain
higher reward. None of the random sets Pareto-dominates the learned set
(Figure~\ref{fig:random50}). The learned set therefore supplies a
preservation-oriented operating point within the sampled trade-off. This is a comparison against a distribution of fixed random sets. A search comparison would choose candidates on optimization data under a common rollout budget, then evaluate the chosen sets separately.

Per-image metric records for the learned reference are unavailable in the audited log. Random-set standard deviations describe variation across fixed sets on these images, not uncertainty across images or training seeds. This comparison does not establish that policy optimization outperforms random search under a matched search budget.

\section{End-to-End Evaluation Details}
\label{sec:supp_end_to_end}

All methods use the same 60 validation examples, prompts, generation seed
42, and target guidance 3.5. Metrics use $512\times512$ outputs. Stable
Flow generates at $1024\times1024$ and is bicubically downsampled before
evaluation. Table~\ref{tab:end_to_end} reports the metrics, and
Table~\ref{tab:supp_method_configs} lists the solver and injection settings.

\begin{table}[t]
    \centering
    \small
    \setlength{\tabcolsep}{4pt}
    \renewcommand{\arraystretch}{1.06}
    \begin{tabular}{@{}lp{0.66\columnwidth}@{}}
        \toprule
        Method & Configuration \\
        \midrule
        RF-Edit & 25/25, $N=5$; 18 value blocks; no mask \\
        FireFlow & 8-step, $N=1$; 38 value blocks; no spatial mask \\
        DNAEdit & 24/24; no injection \\
        ReFlex & 21/28; 32 sharing blocks; native mask \\
        Stable Flow & 50/50; 9 K/V-sharing blocks; no spatial mask \\
        Ours (RF-Edit) & 25/25, $N=5$; 18 output blocks;
        endpoint mask, $\rho=.2$ \\
        Ours (learned) & 25/25, $N=5$; 9 output blocks;
        endpoint mask, $\rho=.2$ \\
        \bottomrule
    \end{tabular}
    \caption{End-to-end configurations. Baselines retain their native solver
    and injection configurations; ours use RF-Solver; slashes separate source and target
    steps. All methods use target guidance 3.5; our RF-Solver runs use
    inversion guidance 1.0.}
    \label{tab:supp_method_configs}
\end{table}

\paragraph{Run accounting.}
Each deterministic point is one pass over its stated split with seed 42.
The controlled 140-example layer-set and mask comparisons use three fixed
random draws each; the separate validation block-set comparison uses 50
fixed random sets and 3,000 random-set edits. Every
objective--budget configuration is trained once and evaluated for each
reported $N$; the step-160 policy continues its step-80 run.

\begin{figure*}[p]
    \centering
    \includegraphics[width=\textwidth]{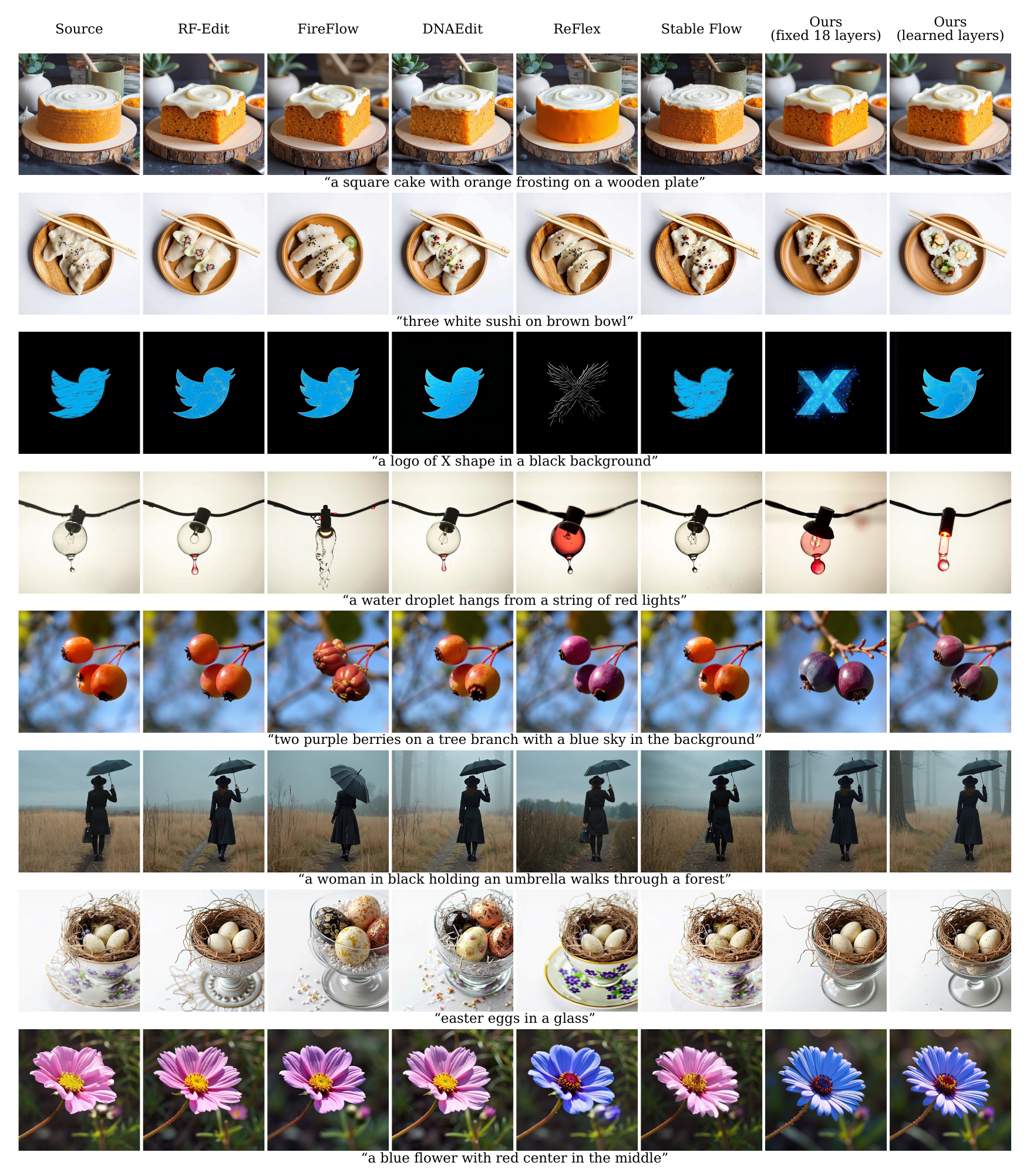}
    \caption{Qualitative comparison from the 140-example analysis category. Rows show the source, five
    baselines, and our fixed 18-layer and learned Top-$9$ results; prompts
    appear below.}
    \label{fig:supp_qualitative_comparison}
\end{figure*}

\paragraph{Qualitative comparison.}
Figure~\ref{fig:supp_qualitative_comparison} shows examples from the
separate 140-example analysis runs, not from the 60-example quantitative
validation evaluation.

\end{document}